\documentclass[runningheads]{llncs}

\pdfoutput=1

\usepackage{color}
\usepackage{xspace}
\usepackage{xcolor}
\usepackage{booktabs}
\usepackage{graphicx}
\usepackage{graphicx}
\usepackage{breakcites}
\usepackage[T1]{fontenc}
\usepackage{amsmath,amssymb,amsfonts}
\usepackage[breaklinks=true]{hyperref}

\newcommand{\ours}{Dif4FF\xspace}

\begin{document}

\title{Dif4FF: Leveraging Multimodal Diffusion Models and Graph Neural Networks for Accurate New Fashion Product Performance Forecasting}

\titlerunning{Dif4FF: Diffusion Models for New Fashion Product Forecasting}

\author{
Andrea Avogaro\inst{1} \and
Luigi Capogrosso\inst{1} \and
Franco Fummi\inst{1} \and
Marco Cristani\inst{1,2}
}
\authorrunning{A.~Avogaro et al.}

\institute{Dept. of Engineering for Innovation Medicine, University of Verona, Verona, Italy
\email{name.surname@univr.it} \and
QUALYCO S.r.l., Spin-off of the University of Verona, Verona, Italy
\email{marco.cristani@qualyco.com}
}

\maketitle

\begin{abstract}
In the fast-fashion industry, overproduction and unsold inventory create significant environmental problems.
Precise sales forecasts for unreleased items could drastically improve the efficiency and profits of industries.
However, predicting the success of entirely new styles is difficult due to the absence of past data and ever-changing trends.
Specifically, currently used deterministic models struggle with domain shifts when encountering items outside their training data.
The recently proposed diffusion models address this issue using a continuous-time diffusion process.
Specifically, these models enable us to predict the sales of new items, mitigating the domain shift challenges encountered by deterministic models.
As a result, this paper proposes Dif4FF, a novel two-stage pipeline for New Fashion Product Performance Forecasting (NFPPF) that leverages the power of diffusion models conditioned on multimodal data related to specific clothes.
Dif4FF first utilizes a multimodal score-based diffusion model to forecast multiple sales trajectories for various garments over time.
The forecasts are refined using a powerful Graph Convolutional Network (GCN) architecture.
By leveraging the GCN's capability to capture long-range dependencies within both the temporal and spatial data and seeking the optimal solution between these two dimensions, Dif4FF offers the most accurate and efficient forecasting system available in the literature for predicting the sales of new items.
We tested Dif4FF on VISUELLE, the de facto standard for NFPPF, achieving new state-of-the-art results.


\end{abstract}

\section{Introduction} \label{sec:sec_intro}

The fast fashion industry's resource demand and reliance on disposable products have led to a crisis of pollution and waste~\cite{niinimaki2020environmental}.
From the vast quantities of water consumed to the mountains of textile waste generated, the industry's impact on the planet is undeniable~\cite{bailey2022environmental}.

Predicting sales for new products more accurately could be a game-changer for this industry.
Reducing overproduction and minimizing waste can make fast fashion more sustainable and less environmentally harmful.
While forecasting historical trends is relatively straightforward, predicting new products' future is a more complex challenge.

Accurate predictions can offer significant benefits. Not only can they help us protect the environment, but they can also lead to substantial economic savings.
By reducing waste and optimizing production, companies can improve their bottom line.
Additionally, pursuing accurate predictions can drive innovation towards more sustainable production methods.

This problem, known as \textit{New Fashion Product Performance Forecasting (NFPPF)}~\cite{skenderi2024well}, is a challenging task due to the absence of historical sales data.
To address this, researchers must leverage available information, such as product specifications (color, type, material), release timing, and past interest in similar items~\cite{skenderi2024well}.
In this direction, Advanced deep learning methods can potentially extract valuable insights from these data points to create accurate forecasts for new fashion products.

\paragraph*{\textbf{Motivation for this paper.}}
Given the industry's fast-paced trends, the definition of what is fashionable and what is not changes rapidly (\emph{i.e.}, in years or even seasons), making it difficult to understand precisely what the market performance could be for a specific item and which are the factors worth considering.
To date, ``classical'' deterministic forecast models have shown quasi-satisfactory performance in certain scenarios.
Still, they have a drawback: they assume that the same characteristics of past season items are also applicable to new items, which, in reality, are often different.
Indeed, this leads to unrealistic predictions due to the shift in input features' domain.

\begin{figure*}[t!]
    \centering
    \includegraphics[width=\linewidth]{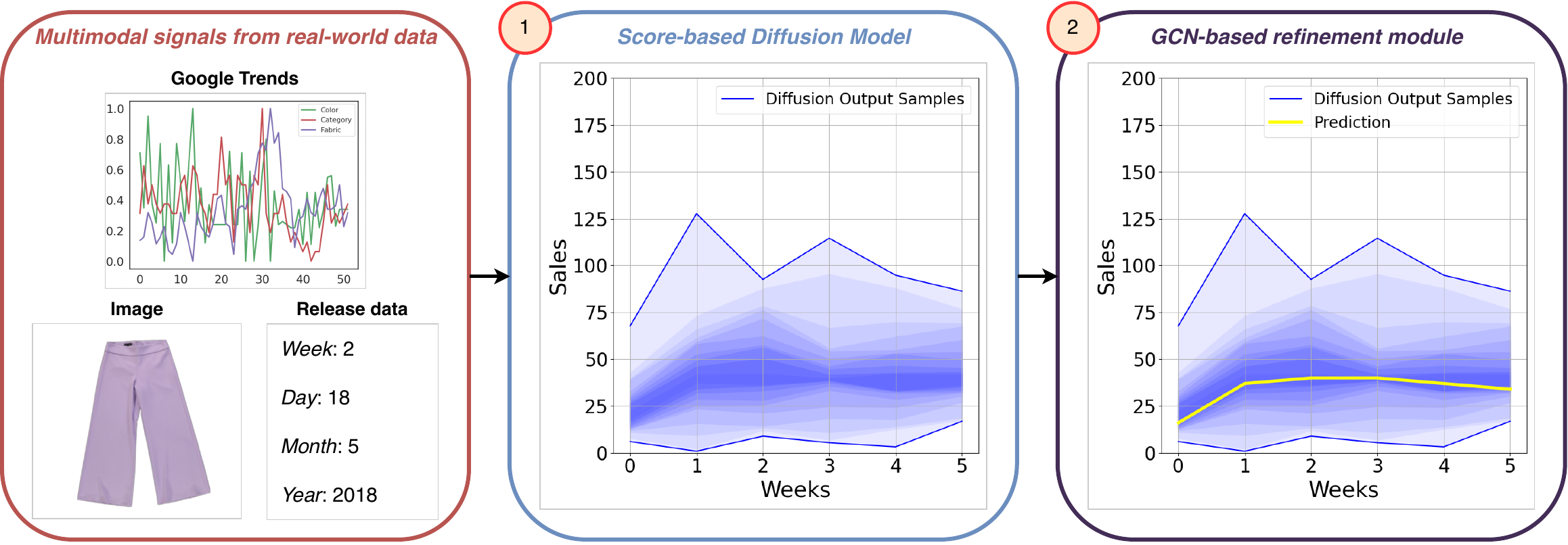}
    \caption{
    \ours{}: a two-stage pipeline for New Fashion Product Performance Forecasting (NFPPF).
    Starting from multimodal signals (\emph{i.e.}, the image, the release data, and the Google Trends) of a single fashion product, we build a multimodal score-based diffusion model to generate an initial prediction of the sales, addressing potential objects with features beyond the training distribution.
    Then, we refine the Diffusion output using a powerful Graph Convolutional Network (GCN) architecture to obtain the final prediction.}
    \label{fig:fig_figure_1}
\end{figure*}
In the field of data modeling, diffusion-based approaches, exemplified by Denoising Diffusion Probabilistic Models DDPMs~\cite{ho2020denoising}, have emerged as powerful tools.
Their ability to capture the underlying probabilistic structure of data has proven invaluable in tasks ranging from visual content generation to time series forecasting~\cite{lin2023diffusion}.
Particularly in the context of fashion sales forecasting, where data features can fluctuate rapidly, DDPMs show a remarkable ability to adapt and maintain accurate predictions, even when faced with patterns of new or unknown features.
Unlike traditional deterministic models that rely on direct mappings between input features and output predictions, DDPMs exploit a diffusion process that gradually refines the noisy data, ensuring that predictions remain grounded in the observed data distribution~\cite{sohl2015deep}.
This adaptive nature makes DDPMs a compelling choice for forecasting challenges characterized by non-stationarity and changing trends.

When encountering previously unseen fashion trends, diffusion models like DDPMs exhibit a unique behavior.
Their predictions stay within the boundaries of what they've learned from their training data\cite{yang2023diffusion}, and his property is desirable, especially in a rapidly changing industry like fast fashion.
Moreover, these models don't stray too far from the established patterns even when provided with additional information about specific features.
This contrasts with deterministic methods, which directly connect input features to sales predictions.
Such a direct approach can be risky, as it might lead to inaccurate forecasts for novel combinations of features.

\paragraph*{\textbf{Innovations in this paper.}}
As a result, in this study, we introduce \ours{}: the first two-step multimodal Diffusion-based pipeline for NFPPF, shown in Figure~\ref{fig:fig_figure_1}.
Firstly, we build and train a multimodal score-based diffusion model, \emph{i.e.}, a generalization of a DDPM, to provide initial predictions and handle cases with features beyond the training distribution.
Secondly, we refine the diffusion model outputs using a powerful Graph Convolutional Network (GCN) architecture.
Despite the effectiveness of diffusion models, their output is aleatory and can strongly differ from that of other models that generate outcomes starting from the same conditioning.
For this reason, we generate multiple predictions for each object, exactly 50, to be precise, and then use them as inputs for the GCN model to have the final prediction.
This strategy ensures smoother in-distribution data as input for the GCN, better reflecting the sales data distribution, thus enhancing the pipeline's reliability.
Specifically, our GCN builds two types of graphs from the input.
One graph focuses on the time dimension, highlighting important connections among the weak sales.
Then, we create another graph based on prediction space, pinpointing strong connections among model-predicted samples.
Finally, we use three Conv1D layers to compress the graph network's output and generate the prediction vector.
We evaluated \ours{} on VISUELLE~\cite{skenderi2022multi,skenderi2024well}, the leading benchmark for NFPPF, and achieved state-of-the-art results.

To summarize, the main contributions of our work are as follows:
\begin{itemize}
\item We propose \ours{}, the first two-stage pipeline for NFPPF based on multimodal diffusion models.
\item In particular, we first build and train a multimodal score-based diffusion model to provide initial predictions, handling cases with features beyond the training distribution.
\item Secondly, we refine the diffusion model outputs through a powerful GCN model to have the final prediction.
\item We tested it on VISUELLE, the de facto standard for NFPPF, achieving state-of-the-art results.
\end{itemize}

The rest of the paper is organized as follows.
Section~\ref{sec:sec_related} presents background information.
Our proposal's details are outlined in Section~\ref{sec:sec_method}, followed by experiments in Section~\ref{sec:sec_experiments}.
Finally, conclusions are drawn in Section~\ref{sec:sec_conclusions}.

\section{Related Works} \label{sec:sec_related}

\paragraph*{\textbf{NFPPF problem.}} \label{subsec:subsec_nfppf_problem}
The existing literature on NFPPF with deep learning tools is limited but growing.
One of the first papers that investigates the sales forecasting problem is~\cite {ren2017comparative}.
Following,~\cite{craparotta2019siamese} attempts to emulate an expert's decision-making process, training a ConvNet-based model to extract visual features from the image and then, through the k-Nearest Neighbors (k-NN) algorithm, confront the features with other elements already seen to produce the final prediction of the sales.

With~\cite{singh2019fashion,ekambaram2020attention}, there was a first try at tackling this task by exploring various algorithms.
In particular,~\cite{singh2019fashion} compares several machine learning algorithms, like gradient boosting and random forest, discussing which is better for NFPPF.
Moreover, the authors also tried two deep learning approaches, \emph{i.e.}, Feed-Forward Networks (FFNs) and Long Short-Term Memory (LSTM), both fed with multimodal signals.
Those signals were obtained from static attributes of the items like category, color, fabric, and variable information such as discounts and promotions.
Similarly,~\cite{ekambaram2020attention} uses an architecture based on Recurrent Neural Networks (RNNs), including more signals like past sales, images, textual embeddings, and discounts.
The model also operates a soft-attention mechanism to understand which information is more relevant to produce the forecasted sales signal.
However, their autoregressive model produced the same prediction across products of different seasons.
Unfortunately, the authors' code and dataset are proprietary and not publicly unavailable.

Building on the value of exogenous signals in fashion forecasting,~\cite{skenderi2024well} propose an encoder-decoder Transformer-based architecture incorporating as input of the model all the multimodal data offered by the dataset~\cite{skenderi2022multi}.
The encoder is fed with Google Trend signals, while the decoder receives an ensemble of features extracted from images, textual descriptions, and the item's temporal information (release date).
This approach effectively extends previous works by leveraging a more powerful architecture to extract insights from exogenous signals, showing the effectiveness of the Google Trends information regarding NFPPF.

In contrast to prior approaches, this paper introduces the first diffusion model-based implementation for solving the NFPPF task.
In this way, we solve the problem common to all the previous methods: unrealistic predictions due to the shift in input features' domain.

\paragraph*{\textbf{Datasets for NFPPF.}}
Publicly available datasets for fashion forecasting, like~\cite{hmdataset}, take into account diverse applications that are dissimilar from NFPPF.
They have usually been used to forecast fashion styles, which are aggregates of products of multiple brands in terms of popularity based on social media, such as, for example, Instagram.
In our case, the task is different since we focus only on single products and not on groups of products, so we have less data to reason on. 
In addition, we are considering genuine sales data and not popularity trends.
As a result, in our paper, we use the VISUELLE dataset~\cite{skenderi2022multi,skenderi2024well}, the only dataset available in the literature for NFPPF, and so the de facto standard for this task.
Due to its nature, our research is also impactful from an industrial level.

\section{Methodology} \label{sec:sec_method}

In this section, we discuss the problem formalization (Section~\ref{subsec:subsec_problem_formalization}).
Then, we delve into the theoretical background of diffusion-based generation models (Section~\ref{subsec:subsec_background}).
Next, we introduce our novel multimodal score-based diffusion model (Section~\ref{subsec:subsec_score_dm}).
Following that, we discuss how we guide the generation process using multimodal information (Section~\ref{subsec:subsec_multimodal_conditioning}).
Finally, we explore the graph neural network-based refinement (Section~\ref{subsec:subsec_refinement_module}).

\subsection{Problem Formalization} \label{subsec:subsec_problem_formalization}
Given a new product $j$, we want to predict $y\in\mathbb{R}^{W}$ expressed as the performance vector in terms of sales in an interval of $W$ weeks since its release date.

For every $j$, a set of three attributes is given: an image of the product $i_j \in \mathbb{R}^{w\times h\times 3}$ with $w=h=256$ and release date $t_j\in\mathbb{R}^4$ composed by four digits representing day, week, month and year of release.

The last information given is a Google Trends signal~\cite{skenderi2024well} $g_j\in\mathbb{R}^{3 \times 52}$ related to a specific product in a particular interval of time, extracted using as a query the fabric, color, and category of the item on the 52 weeks preceding the release date.

\subsection{Diffusion-Based Generation} \label{subsec:subsec_background}
DDPMs~\cite{sohl2015deep,ho2020denoising} are a class of deep latent variable models that work by modeling the joint distribution of the data over a Markovian inference process.
This process consists of small perturbations of the data with a variance-preserving property~\cite{song2020score}, such that the limit distribution after the diffusion process is approximately identical to a known prior distribution.
Starting with samples from the prior, a reverse diffusion process is learned by gradually denoising the sample to resemble the initial data by the end of the procedure.

Formally, the data distribution $q(x_0)$ is modelled through a latent variable model $p_\theta(x_0)$:
\begin{gather}
    p_\theta(x_0) = \int p_\theta(x_{0:T}) dx_{1:T}\;,\\ \quad  p_\theta(x_{0:T}) := p_\theta(x_T) \prod_{t=1}^{T} p^{(t)}_\theta(x_{t-1} | x_t)\;,
\end{gather}
where $x_1, \ldots, x_T$ are latent variables of the same dimensionality as $x_0$.

The parameters $\theta$ are learned by maximizing an ELBO of the log evidence, \emph{i.e.}:
\begin{gather} \label{eq:eq_elbo}
    \max_\theta \mathbb{E}_{q(x_0)}[\log p_\theta(x_0)] \leq \nonumber \\
    \max_\theta \mathbb{E}_{q(x_0, x_1, \ldots, x_T)}\left[\log p_\theta(x_{0:T}) - \log q(x_{1:T} | x_0) \right]\;,
\end{gather}
where $q(x_{1:T} | x_0)$ represents a fixed inference process defined as a Markov chain:
\begin{gather}
    q(x_{1:T} | x_0) := \prod_{t=1}^{T} q(x_t | x_{t-1})\;,\\
    q(x_t | x_{t-1}) := \mathcal{N}\left(\sqrt{\frac{\alpha_t}{\alpha_{t-1}}} x_{t-1}, \left(1 - \frac{\alpha_t}{\alpha_{t-1}}\right) I\right)\;,
\end{gather}
where $\alpha_{1:T} \in (0, 1]^T$ is a predefined variance schedule, and the covariance matrix is ensured to have positive terms on its diagonal. 
Specifically, this parametrization has the property:
\begin{align}
    q(x_t | x_0) = \int q(x_{1:t} | x_0) dx_{1:(t-1)} = \nonumber \\
    \mathcal{N}(x_t; \sqrt{\alpha_t} x_0, (1 - \alpha_t)I)\;,
\end{align}
therefore we can write $x_t$ as a linear combination of $x_0$ and a noise variable $\epsilon$. 

When we set $\alpha_{T}$ sufficiently close to $0$, $q(x_T | x_0)$ converges to a standard Gaussian for all $x_0$, so it is natural to set $p_\theta(x_T) := \mathcal{N}(0, \mathbf{I})$.
Given that all the conditionals are modeled as Gaussians with fixed variance, the objective in Equation~\ref{eq:eq_elbo} can be greatly simplified. 
In particular,~\cite{ho2020denoising} shows that the following (further simplified) lower bound provides optimal generative performance:
\begin{gather}
    L(\epsilon_\theta) := \sum_{t=1}^{T} \mathbb{E}_{x_0,\epsilon_t}\left[ \lVert {\epsilon_{\theta}^{(t)}(\sqrt{\alpha_t} x_0 + \sqrt{1 - \alpha_t} \epsilon_t) - \epsilon_t}\rVert_2^2 \right]\;,
\end{gather}
where $x_0\sim q(x_0), \epsilon_t \sim \mathcal{N}(0, I)$, $\epsilon_\theta = \{\epsilon_\theta^{(t)}\}_{t=1}^{T}$ is a set of $T$ functions, with each $\epsilon_{\theta}^{(t)}: X \to X$ having trainable parameters $\theta^{(t)}$.

In practice, these functions are approximated by a neural network conditioned on the diffusion time $t$. 
After the model is trained, we can generate new samples by first sampling $x_T$ from the known prior $p_\theta(x_T)$, and then iteratively reversing the diffusion process, thereby sampling $\{x_{T-1}\ldots x_0 \}$.

In addition, we leveraged the natural ability of DDPMs to incorporate multimodal conditioning in the generation process, taking inspiration from~\cite{ho2022classifier,rombach2022high,zhang2023adding,capogrosso2023neuro}.

\subsection{Our Multimodal Score-Based Diffusion Model} \label{subsec:subsec_score_dm}
Score-based diffusion models~\cite{song2020score} generalize DDPMs~\cite{ho2020denoising} generative models trained to reverse a discrete-time diffusion process.
A Gaussian noise diffusion process, also known as the \textit{forward process}, can be summarized as a chain of steps in which Gaussian noise is progressively added to the initial distribution, as described by the following equations:
\begin{equation}
    q(x^{1},...,x^{T}|x^{0}=y)=\prod_{t=1}^{T}q(x^{t}|x^{t-1})\;,
\end{equation}
\begin{equation}
    q(x^{t}|x^{t-1}):=\mathcal{N}(\sqrt{1-\beta_{t}}x_{t-1},\beta_{t}I)\;,
\end{equation}
where $q(x^T)\approx \mathcal{N}(0,1)$, $y=q(x^0)$ is the true data distribution, $\beta_t$ is the variance of the additive noise, and $t\in [0,T]$ represents the number of noising steps defined a prior. 

A model $p_\theta$ is then trained to reverse the diffusion process by gradually removing noise, also known as the \textit{backward process}, to restore the initial distribution.
Specifically, the backward process is formalized as follows:
\begin{equation}
    \begin{split}
    p_\theta(x^{t-1}|x^t,c) = \mathcal{N}(x^{t-1}; \mu_\theta(x^t,t) &\\  + s\sigma_t^2\nabla_{x^t}\text{log}p(x^t|x^0), \sigma_t^2 I)\;,
    \end{split}
\end{equation}
where $\sigma$ is the variance for each timestep, and $s$ is the parameter that controls the strength of the conditioning.

\begin{figure*}[t!]
    \centering
    \includegraphics[width=\linewidth]{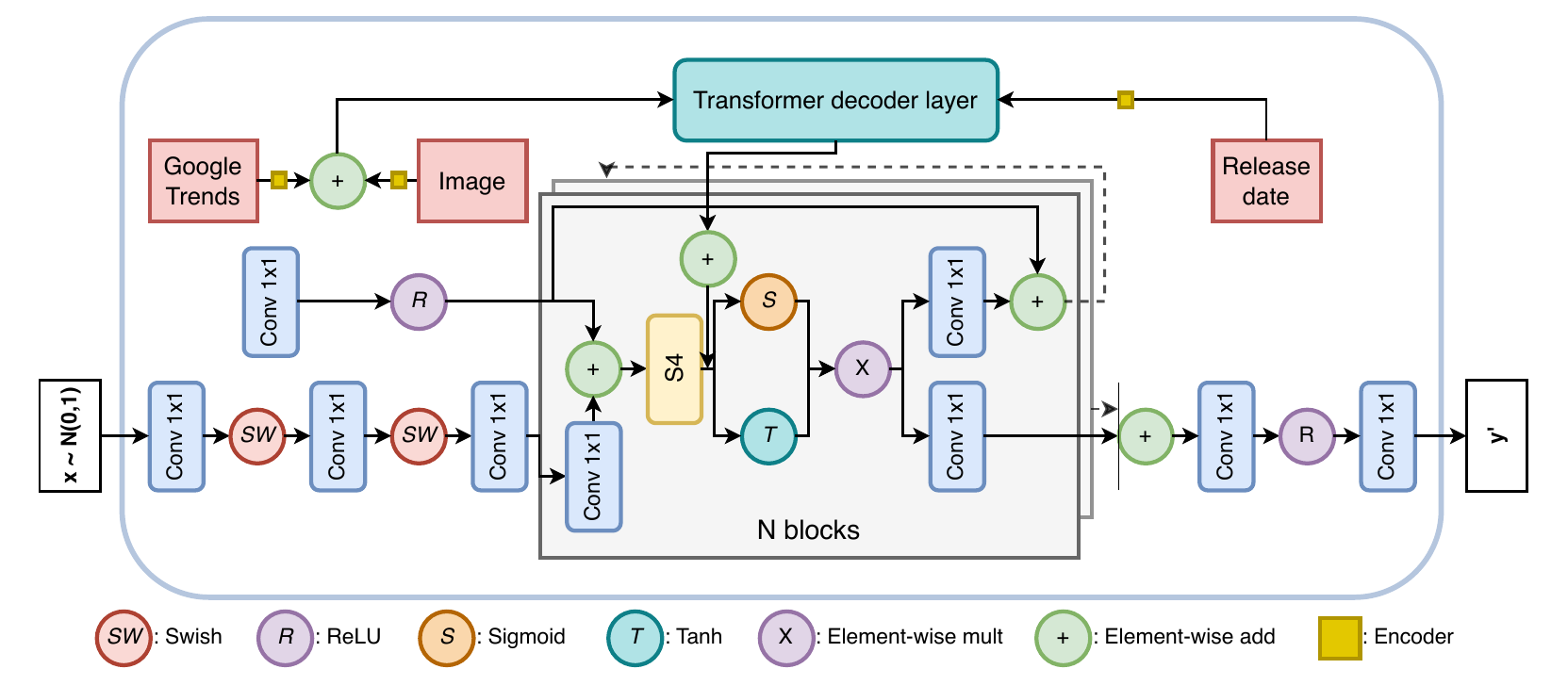}
    \caption{
    An overview of our multimodal score-based diffusion model.
    Each block contains two outputs: one for the subsequent block and another for a skip connection.
    The summation of all skip connections forms the model's final output.
    The primary component of each block is typically an S4 block~\cite{gu2021efficiently}.}
    \label{fig:fig_figure_2}
\end{figure*}
Specifically, Figure~\ref{fig:fig_figure_2} shows the architecture of our multimodal score-based diffusion model.
The network is a stack of multiple $N$ blocks.
Every block has two outputs, one for the next block and one for skip connection.
The summation of all the skip connections represents the actual output of the model.
Every block is mainly composed of an S4 block~\cite{gu2021efficiently}.

\subsection{Multimodal Conditioning} \label{subsec:subsec_multimodal_conditioning}
We leverage a combination of data sources to predict future sales for each product.
This data will include images of the products (represented as $i_j$), the Google Trend signal $g_j$, and their release dates $t_j$.
More details on this process can be found in Section~\ref{subsec:subsec_problem_formalization}.

We train three different encoders $I_{\theta}, T_{\theta}$ and $G_{\theta}$ to extract features from the input information.
Then, these features are used to produce the conditional embedding through a cross-attention mechanism defined as:
\begin{align}
    c_j &= \text{Softmax}\left(\frac{Q_j K_j^T}{\sqrt{d_k}}\right) V_j\;,
\end{align}
where $K_j=V_j=G_{\theta}(g_j) + T_{\theta}(i_j)$, $Q_j=I_{\theta}(i_j)$ and $\sqrt{d_k}$ is the dimensionality of the number of features of the embeddings.

The idea behind this choice is to start from the Google trend signal and use the release date embedding as positional encoding.
Lastly, the Google Trend signal is weighted on the visual features (\emph{i.e.}, shape, color, fabric, etc.) extracted from the image.
The \textit{multimodal conditioning} $c_j$ is directly added to the output of the S4 block, and the cross-attention is implemented using a Transformer decoder layer~\cite{vaswani2017attention}.

\subsection{GCN-based Diffusion Outputs Refinement} \label{subsec:subsec_refinement_module}
The refinement module comprises two main modules: the first is based on ST-GCN~\cite{yan2018spatial}.
The GCN block operates on two different dimensions by constructing two graphs.

The first graph operates on the dimension of the space of the predictions made by the diffusion model.
The mathematical formulation is the following:
\begin{equation}
    X_s = \phi_s(A_s X)\;,
\end{equation}
where $A_s \in \mathbb{R}^{S \times S}$ is the adjacency matrix, $X$ output of the diffusion model and $\phi_s$ an Multi-layer Perceptron (MLP).

The second block works on the time dimension, learning a graph that weighs the connections between the various prediction weeks.
Formally, it is defined as:
\begin{equation}
    X_t = \phi_t(A_t X_s)\;,
\end{equation}
where $A_t \in \mathbb{R}^{W \times W}$ is the adjacency matrix, $X_t$ output of the first GCN block and $\phi_t$ an MLP.

These blocks are designed to compress the $N$ predicted samples and regress the final prediction. 
The network is then trained with a Mean Squared Error (MAE) loss function denoted as $\mathcal{L}_{\text{MAE}}$, mathematically defined as:
\begin{equation}
    \mathcal{L}_{\text{MAE}}(y, \hat{y}) = \frac{1}{N} \sum_{i=1}^{N} |y_i - \hat{y}_i|;,
\end{equation}
where $\hat{y}$ represent the ground truth value and $\hat{y}_i$ is the value predicted by the model.

\section{Experiments} \label{sec:sec_experiments}

Here, we first present the experimental setup (Section~\ref{subsec:subsec_exp_setup}).
Then, we show our quantitative results among eight state-of-the-art competitors (in Section~\ref{subsec:subsec_quantitative}), and we argue regarding the necessity of a refinement module (in Section~\ref{subsec:subsec_qualitative}).
Finally, we report the ablation study in Section~\ref{subsec:subsec_ablation} to further validate our proposal.

\subsection{Experimental Setup} \label{subsec:subsec_exp_setup}

\paragraph*{\textbf{Implementation details.}}
Our model is based on TS-Diff~\cite{kollovieh2024predict}.
This architecture is a score-based diffusion model that is particularly useful for predicting, reconstructing, and refining time series tasks.
TS-Diff uses a series of building blocks called ``S4 Blocks''~\cite{gu2021efficiently}, stacked together four times with connections that allow information to flow directly (skip connections).

We've made some modifications to TS-Diff to incorporate the additional information from images and release dates.
As described in the subsection on multimodal conditioning (Section~\ref{subsec:subsec_multimodal_conditioning}), we've added a layer inspired by Transformers~\cite{vaswani2017attention}.
This layer helps combine the features extracted from images $i_j$, release dates $t_j$, and the Google Trends signals $g_j$ to properly guide the diffusion model.

\textbf{Image encoder.}
We used as Image Encoder $I_{\theta}$ a ResNet-18~\cite{he2016deep} pre-trained on ImageNet-1K~\cite{deng2009imagenet}.
We substituted the last two layers of the model with a Conv1D and a Linear to reduce the dimensionality of the features extracted, obtaining a tensor $I_{\theta}(i_j) \in \mathbb{R}^{C \times W}$, with $C$ channels equal to 64 and $W$ forecasting horizon of six weeks.

\textbf{Temporal encoder.}
The Temporal Encoder $T_{\theta}$ comprises four different MLPs that expand the dimension from 1 to $C$.
The output of this model is then concatenated along the channel dimension and fed into another MLP that reduces the feature number from $4C$ to $C$, resulting in $T_{\theta}(t_j) \in \mathbb{R}^{C}$.

\textbf{Google Trends encoder.}
Lastly, as $G_{\theta}$, we adopted a Transformer Encoder layer~\cite{vaswani2017attention} that performs self-attention operation on the Google Trends.
The encoder layer also reduces the dimensionality of the encoding, resulting in $G_{\theta}(g_j) \in \mathbb{R}^{C\times W}$.

\textbf{GCN-based refinement.}
As described in Section~\ref{subsec:subsec_refinement_module}, the architecture used for the refinement network is based on ST-GCN~\cite{yan2018spatial}.
The first GCN-based module comprises two ST-GCN blocks, with an expansion and a subsequent reduction of the channels to create a first embedding.
On the other hand, the second module is composed of 1D convolutions, reducing the sample dimension from 50 to 1 (to obtain an actual final prediction).
Specifically, this compression is done by three layers of 1D convolutions with PReLUs as activation functions.

The source code is available at \url{https://github.com/andreaavo9/Dif4FF}.

\paragraph*{\textbf{Dataset description.}}
We assessed our model's performance using the VISUELLE fast-fashion dataset~\cite{skenderi2022multi,skenderi2024well}.
This dataset encompasses information on 5,577 products sold across 100 stores of Nunalie, an Italian fast-fashion brand.
Specifically, the experimental protocol simulates how a fast fashion company deals with new products to obtain the best forecast.

We trained our model using multimodal conditioning to simulate a real-world scenario that leverages multiple data sources.
Among the data types provided by the dataset, we used product images, release dates, and Google Trends, skipping the descriptions of each product because they were not informative.
In particular, the Google Trends signals reflect online search interest for products or similar items, potentially indicating consumer demand.
The VISUELLE dataset also includes sales figures for each product across all 100 stores within 12 weeks of its release.

Following the evaluation protocol established in previous studies~\cite{skenderi2022multi,skenderi2024well}, we consider six weeks for the forecast horizon.
All model classifiers have been trained to assume a 12-week prediction, and shorter horizons have been considered for the evaluation.
This procedure maximized the performances of all the approaches, and we used the checkpoint in which the error was the lowest.
To perform the experiments, the data are divided into a training and testing partition, where the testing products are composed of the 497 most recent products.
The rest of the dataset (5,080 products) is used for training.

\paragraph*{\textbf{Evaluation metrics.}}
We used the Mean Average Error (MAE) and Weighted Absolute Percentage Error (WAPE)~\cite{hyndman2008forecasting}, \emph{i.e.}, the two main metrics representing the quality of the forecasting, to evaluate \ours{}.
Formally, they are defined as:
\begin{equation} \label{eq:eq_mae}
    \text{MAE}=\frac{\sum_{t=0}^T |y_t-\hat{y_t}|}{T}\;,
\end{equation}
\begin{equation} \label{eq:eq_wape}
    \text{WAPE}=\frac{\sum_{t=0}^T |y_t-\hat{y_t}|}{\sum_{t=0}^T y_t}\;,
\end{equation}
where $y$ represents the actual values of the time series, $\hat{y}$ represents the forecasted values, and $T$ represents the total number of observations in the time series.

\begin{table*}[t!]
    \centering
    \caption{Quantitative results of \ours{} expressed in terms of WAPE and MAE on VISUELLE, described in Equation~\ref{eq:eq_wape} and Equation~\ref{eq:eq_mae}, respectively.
    In \textbf{bold}, the best results.
    \underline{Underlined}, the second best.}
    \begin{tabular}{l|cccc|cc}
    \toprule
    \textbf{Method} & \textbf{\textsc{Image}} & \textbf{\textsc{Release}} & \textbf{\textsc{Descr.}} & \textbf{\textsc{Google T.}} & \textbf{WAPE $\downarrow$} & \textbf{MAE $\downarrow$} \\
    \midrule
    Mean predictor    & & & & & 60.1 & 32.8 \\
    Median predictor  & & & & & 58.3 & 31.8 \\
    \midrule
    Attribute k-NN~\cite{ekambaram2020attention} & & & \checkmark & & 59.8 & 32.7 \\
    Image k-NN~\cite{ekambaram2020attention} & \checkmark & & & & 62.2 & 34.0 \\
    Attr+Image k-NN~\cite{ekambaram2020attention} & \checkmark & & \checkmark & & 61.3 & 33.5 \\
    GBoosting~\cite{friedman2001greedy} & \checkmark & \checkmark & & & 64.1 & 35.0  \\
    GBoosting+G~\cite{friedman2001greedy} & \checkmark & \checkmark & & \checkmark & 63.5 & 34.7 \\
    Cat-MM-RNN~\cite{ekambaram2020attention} & & \checkmark & \checkmark & \checkmark & 63.3 & 34.4\\
    X-Att-RNN~\cite{ekambaram2020attention} & \checkmark & \checkmark & \checkmark & & 59.5 & 32.3 \\
    GTM-Transformer~\cite{skenderi2024well} & \checkmark & \checkmark & \checkmark & \checkmark & \underline{55.2} & \underline{30.2} \\
    \midrule
    \ours{} (ours) & \checkmark & \checkmark & & \checkmark & \textbf{54.6} & \textbf{30.0} \\
    \bottomrule
    \end{tabular}
    \label{tab:tab_table_1}
\end{table*}

\paragraph*{\textbf{Training details.}}
All the code is implemented in PyTorch~\cite{paszke2019pytorch}.
For the multimodal score-based diffusion model, we train the network for 500 epochs, with a learning rate of $1\times{}10^{-3}$, a weight-decay of $5\times{}10^{-4}$, using AdamW~\cite{loshchilov2017decoupled} as an optimizer, on a NVIDIA RTX 4090.

Given the nature of diffusion models, different seeds produce different predictions.
As a result, we ensured that our score-based diffusion model was executed in the most deterministic setup possible by setting the seed value to 32 across all libraries used.
A Bayesian algorithm was used for the GCN networks to search for the best training hyperparameters.

\subsection{Quantitative Results} \label{subsec:subsec_quantitative}
Here, we discuss the quantitative results obtained with \ours{}.
As we can see from Table~\ref{tab:tab_table_1}, \ours{} outperforms all the other state-of-the-art methods.
To verify the statistical significance of the results, we ran 10 instances of the \ours{} pipeline.
The mean MAE was 30.2 with a variance of 0.04, and the mean WAPE was 54.8 with a variance of 0.08.

In \ours{}, we don't use the information from the textual description of the various samples since these could worsen the model's performance in our diffusion-based architecture.
Furthermore, the diffusion model probably retrieves the color, fabric, and color information directly from the image features.
Furthermore, an objective product description is notoriously difficult to obtain in a real-world scenario since fashion garments description involves a complex interplay of abstract concepts and stylistic elements, making objective descriptions for each garment in the dataset challenging~\cite{shimizu2023fashion}.
For more information about that, see Section~\ref{subsec:subsec_ablation}.

In particular, we report the comparisons among eight other existing models.
\cite{skenderi2024well} is the most similar in terms of performance to \ours{}.
The most noticeable improvement can be seen in WAPE, which has a sharp drop from 55.2 to 54.7, with a slight improvement in MAE as well.
These enhancements in performance offer significant benefits that might not be immediately apparent.
Primarily, as already introduced in Section~\ref{sec:sec_intro}, a diffusion-based model is always preferable for this task since it should better maintain performance with out-of-distribution objects, ensuring greater stability in practice when used in real-world contexts.
Secondly, given our specific architecture, we need less information to achieve better results with respect to the current state-of-the-art.

\subsection{The Diffusion Outputs Refinement} \label{subsec:subsec_qualitative}
\begin{figure*}[!t]
    \centering
    \includegraphics[width=\linewidth]{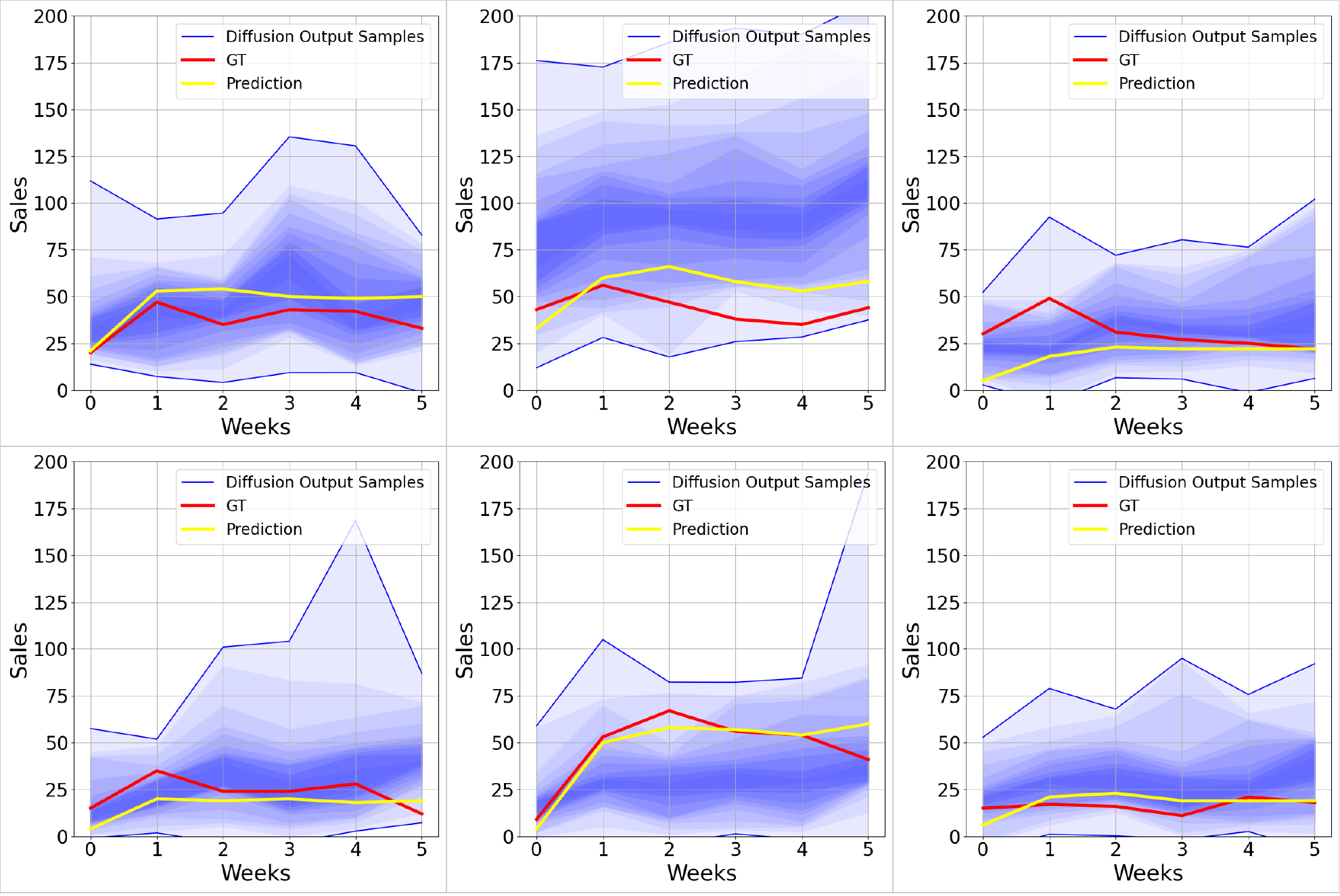}
    \caption{In the figures above are presented some visual representations of the multimodal score-based diffusion model outputs.
    In particular, the blue region represents the output distribution of the diffusion model given a certain sample.
    Specifically, the blue area is obtained by computing the weekly quantiles among the 50 outputs.
    The Prediction line, on the other hand, is the output of the refinement GCN, \emph{i.e.}, the final prediction.
    The forecasting period is six weeks from the release date, depicted on the x-axis.
    On the y-axis, the number of units sold of a specific garment in the various shops is shown.}
    \label{fig:fig_figure_3}
\end{figure*}

In this section, we explain the role of the GCN in refining the multimodal score-based diffusion model output forecasting sales.
Starting with the visualization of a few samples in Figure~\ref{fig:fig_figure_3}, it can be seen that the model itself gives as output a distribution of predictions that follows the ground truth very closely.
Therefore, the role of the refinement GCN may seem obsolete, but there are several factors to consider that make it crucial.

Firstly, the model output consists of $N$ predictions; it is then necessary to use some technique to obtain a single final prediction.
Examples might be simply taking the predictions' mean or median.
However, this would result in performance degradation as the ground truth often differs from the distribution's mean or median. 

As a result, we implemented a powerful GCN network trained to refine the diffusion model output, as described in Section~\ref{subsec:subsec_refinement_module}.
As we can see from Figure~\ref{fig:fig_figure_3}, specifically in the second and fifth images, the ground truth does not reside in the densest region of the distribution, and the refinement network very effectively follows its movement away from the median of the diffusion model outputs.

In order to understand how much the refinement model actually helps to improve the performance (and not just predict the mean/median) of the diffusion distribution output, the errors in terms of MAE and WAPE of both static measurements were calculated.
Using the median to extract a final prediction, we obtained a result of 34.8 of MAE and 59.3 of WAPE.
On the other hand, for the mean, an MAE error of 34.5 and WAPE of 58.7 were obtained. 
Thus, adopting a refinement network is a winning strategy, improving performance considerably without excessively increasing the model complexity.

\subsection{Ablation Studies} \label{subsec:subsec_ablation}
\begin{table*}[t!]
    \centering
    \caption{Table representing the different tests made with the same multimodal score-based diffusion model.
    We tested our model first without the temporal condition and then without images.}
    \begin{tabular}{l|ccc}
    \toprule
    \textbf{Model} & \textbf{WAPE $\downarrow$} & \textbf{MAE $\downarrow$} \\
    \midrule
    \ours{} (ours) without the temporal condition  & 56.6 & 31.5 \\
    \ours{} (ours) without the images              & 56.2 & 31.4 \\
    \ours{} (ours) without the Google Trends       & 55.6 & 30.9 \\
    \ours{} (ours) & \textbf{54.6} & \textbf{30.0} \\
    \bottomrule
    \end{tabular}
    \label{tab:tab_table_2}
\end{table*}

\begin{table}[t!]
    \centering
    \caption{Table representing the results obtained in the domain-shift example.
    It is clear that our diffusion model is more resilient to the domain shift due to the different years it has been used compared to the second-best performing method.}
    \begin{tabular}{l|ccc}
    \toprule
    \textbf{Model} & \textbf{WAPE $\downarrow$} & \textbf{MAE $\downarrow$} \\
    \midrule
    GTM-Transformer \cite{skenderi2024well} & 56.9 & 32.8 \\
    \ours{} (ours) &\textbf{55.9} &\textbf{31.4} \\
    \bottomrule
    \end{tabular}
    \label{tab:tab_table_3}
\end{table}

The first ablation study tests the model's performance using different conditioning setups.
It should be noted that the error values of each test were obtained by running the entire \ours{} pipeline.
As we can see from Table~\ref{tab:tab_table_2} it is clear that images and temporal information are crucial for the model to predict sales accurately.
This is because the item's features without information on the season and the release period are insufficient to predict an accurate sales value.
Since the fashion market is a sector strongly influenced by trends, a certain garment may be very fashionable in one season but remain completely unsold in the next.
On the other hand, it is quite straightforward to understand why temporal information and Google Trends, without any further details on the item's color, fabric, or shape, are insufficient to determine an accurate prediction of sales.
Google Trends represents information about the public appreciation of a certain item, and the impact on performance is lower when removed.

The second ablation study is related to the domain shift, a well-known issue in the fashion domain.
To check the resiliency of the models to this phenomenon, we trained both GTM-Transformer and \ours{}, removing from the train set every garment related to 2018, leaving all of them just in the test set.
The test set of VISUELLE is already composed of only 2018 garments.
As shown in Table~\ref{tab:tab_table_3}, our method significantly reduces the error when tested on garments entirely outside the training distribution.
This highlights that, while existing methods may perform adequately on training data, their real-world performance suffers significantly under domain shift.
As a result, \ours{} becomes a killer application for real-world scenarios where domain shift is a common challenge.

Finally, since it is well-recognized that one common issue with diffusion models is their tendency to converge on the mean of the data distribution, we investigated the performance of \ours{} compared to a simple predictor that uses only the mean and variance of the training data for the test set.
The results are reported in Table~\ref{tab:tab_table_1} and reveal that based on the mean and variance, this naive predictor significantly underperforms compared to our \ours{}.
This finding reinforces the effectiveness of our proposed approach for the NFPPF task.

\section{Conclusions} \label{sec:sec_conclusions}

In this paper, we propose \ours{}, a novel two-step multimodal diffusion-based pipeline for NFPPF.
In particular, we first build and train a multimodal score-based diffusion model to provide initial predictions, handling cases with features beyond the training distribution.
Despite the effectiveness of diffusion models, they can sometimes produce inconsistent predictions for the same object.
To tackle this, we generate multiple predictions for each sample using the diffusion model and then use them as inputs for the Graph Convolutional Network (GCN) model to make the final prediction.
We tested \ours{} on VISUELLE, the most widely used benchmark for NFPPF, achieving state-of-the-art results.
As a result, this paper further encourages the research of diffusion models for NFPPF.

\paragraph*{\textbf{Future works.}}
We aim to extend our research by exploring the integration of additional data sources, such as customer feedback, to further enhance the prediction accuracy of \ours{}.
Furthermore, we intend to conduct real-world experiments in collaboration with industry partners to validate the effectiveness of our approach in practical settings.
Finally, a more long-term research line is to move from a two-stage pipeline to an end-to-end system.

\section*{Acknowledgement}
This study was carried out within the PNRR research activities of the consortium iNEST (Interconnected North-Est Innovation Ecosystem) funded by the European Union Next-GenerationEU (Piano Nazionale di Ripresa e Resilienza (PNRR) – Missione 4 Componente 2, Investimento 1.5 – D.D. 1058  23/06/2022, ECS\_00000043).
This manuscript reflects only the Authors’ views and opinions.
Neither the European Union nor the European Commission can be considered responsible for them.

\bibliographystyle{splncs04}
\bibliography{01_bibi}

\end{document}